\documentclass{article}

\usepackage{arxiv}
\usepackage{graphicx}
\usepackage{newtxtext,newtxmath}
\usepackage[utf8]{inputenc} % allow utf-8 input
\usepackage[T1]{fontenc}    % use 8-bit T1 fonts
\usepackage{hyperref}       % hyperlinks
\usepackage{url}            % simple URL typesetting
\usepackage{booktabs}       % professional-quality tables
\usepackage{amsfonts}       % blackboard math symbols
\usepackage{nicefrac}       % compact symbols for 1/2, etc.
\usepackage{microtype}      % microtypography
\usepackage{lipsum}

\title{Mixed Pooling Multi-View Attention Autoencoder for Representation Learning in Healthcare}

\author{
  Shaika Chowdhury \\
  University of Illinois at Chicago\\
  \texttt{schowd21@uic.edu} \\
   \And
 Chenwei Zhang \\
  Amazon\\
  \texttt{cwzhang@amazon.com} \\
  \And
   Philip S.Yu \\
  University of Illinois at Chicago\\
  \texttt{psyu@uic.edu} \\
   \And
 Yuan Luo \\
  Northwestern University\\
  \texttt{yuan.luo@northwestern.edu} \\ 
}

\begin{document}
\maketitle

\begin{abstract}
Distributed representations have been used to support downstream tasks in healthcare recently. Healthcare data (e.g., electronic health records) contain multiple modalities of data from heterogeneous sources that can provide complementary information, alongside an added dimension to learning personalized patient representations. To this end, in this paper we propose a novel unsupervised encoder-decoder model, namely Mixed Pooling Multi-View Attention Autoencoder (MPVAA), that generates patient representations encapsulating a holistic view of their medical profile. Specifically, by first learning personalized graph embeddings pertaining to each patient's heterogeneous healthcare data, it then integrates the non-linear relationships among them into a unified representation through multi-view attention mechanism. Additionally, a mixed pooling strategy is incorporated in the encoding step to learn diverse information specific to each data modality. Experiments conducted for multiple tasks demonstrate the effectiveness of the proposed model over the state-of-the-art representation learning methods in healthcare.
\end{abstract}

% keywords can be removed

\section{Introduction}
Distributed representations, also known as embeddings, have brought immense success in numerous Natural Language Processing (NLP) and Computer Vision \cite{frome2013devise} tasks. Recent works in Deep Learning applications for healthcare data (e.g., EHR) \cite{ choi2016medical, tran2015learning} emulated the concept of embedding as learning vector representations of medical concepts, that ensure that similar concepts will form natural clusters and relationships in vector space \cite{shickel2017deep}. Some other works generated visit representations from the learned medical concept embeddings \cite{choi2016multi, choi2017gram}. Considering each patient as a sequence of these visits, patient representation is then learned through the optimization of a supervised learning task. However, large amount of labeled healthcare data for predictive modeling is not available in practice, not to mention that manual labeling is laborious and expensive. 
Aiming towards alleviating annotation efforts, we propose an unsupervised model that relies only on the large unannotated EHR data to generate patient representations. We call this model the Mixed Pooling Multi-View Attention Autoencoder (MPVAA).

In healthcare data (e.g., EHR), patient records may be available as heterogeneous data (e.g., demographics, laboratory results, clinical notes) that can provide an added dimension to learning personalized patient representations. For example, a patient with ``diabetes'' will have different attributes from a patient with ``condition of heart failure'', and can also further vary among patients with different types of ``diabetes'' (e.g., type I, type II). Prior works focused on learning representations from predominantly one data type (e.g., unstructured notes or structured clinical events) that excludes relevant information available in other modalities. For example, a patient's symptoms for a disease can be mentioned in physician notes, but could be missing from their structured clinical event data specified as medical codes. Therefore, to further improve the information usage with heterogeneous data, in this work we treat the different data modalities in EHR as separate views (i.e., inner view) to first learn patient-specific medical concept embeddings with graph autoencoder. Information from the embedding spaces of the different views are then fused together through attention mechanism to learn an unified patient representation. 
 %However, trying to fuse information from different modalities in EHR is difficult: \textbf{(1)} usage of medical concept terminologies is inconsistent between the structured clinical (i.e., ICD-9/ICD-10 codes) and mention of formal medical terms in clinical notes  \textbf{(2)} in unstructured clinical notes, the context is not as well defined as in clinical codes \textbf{(3)} modeling the non-linear and complex relations among different modalities is not straightforward and \textbf{(4)} The resulting learned embedding should be interpretable.

%Attention mechanism \cite{bahdanau2014neural} provides an intuitive way to incorporate the distinct semantic and syntactic properties of the heterogeneous data types and learn meaningful patient representations. 

Our attention autoencoder model, MPVAA, follows the encoder-decoder architecture. On the encoder side, a multi-layer Transformer encoder \cite{vaswani2017attention} is first applied on the input patient vector, followed by a \textit{mixed pooling} strategy that combines mean-max pooling in a stochastic manner. As mean pooling and max pooling methods have their own advantages and drawbacks \cite{yu2014mixed}, randomly weighting their importance results in a latent representation that encapsulates the most salient feature of the patient vector alongside capturing its general features. The decoder then reconstructs the input vector from the encoded representation through \textit{multi-view attention} mechanism to reconcile the interactions from the heterogeneous features. 

Our contributions in this paper can be summarized as follows:
\begin{itemize}
    \item We propose a new architecture MPVAA for learning patient representations, in which the various locally and globally relevant features present in the heterogeneous information associated with the patient are seamlessly integrated by facilitating interactions of cross-modal features. Multi-view graph is constructed for each patient to promote personalization of the learned representation. 
    \item MPVAA is unsupervised and can be easily generalized to other domains with heterogeneous data.
    \item We evaluate MPVAA  on a publicly available EHR dataset on two different tasks. The results demonstrate the effectiveness of the MPVAA  model compared to the other state-of-the-art models. 
\end{itemize}

\section{Mixed Pooling Multi-View Attention Autoencoder}
\begin{figure}[htb!]
    \centering
    \includegraphics[width=0.7\linewidth]{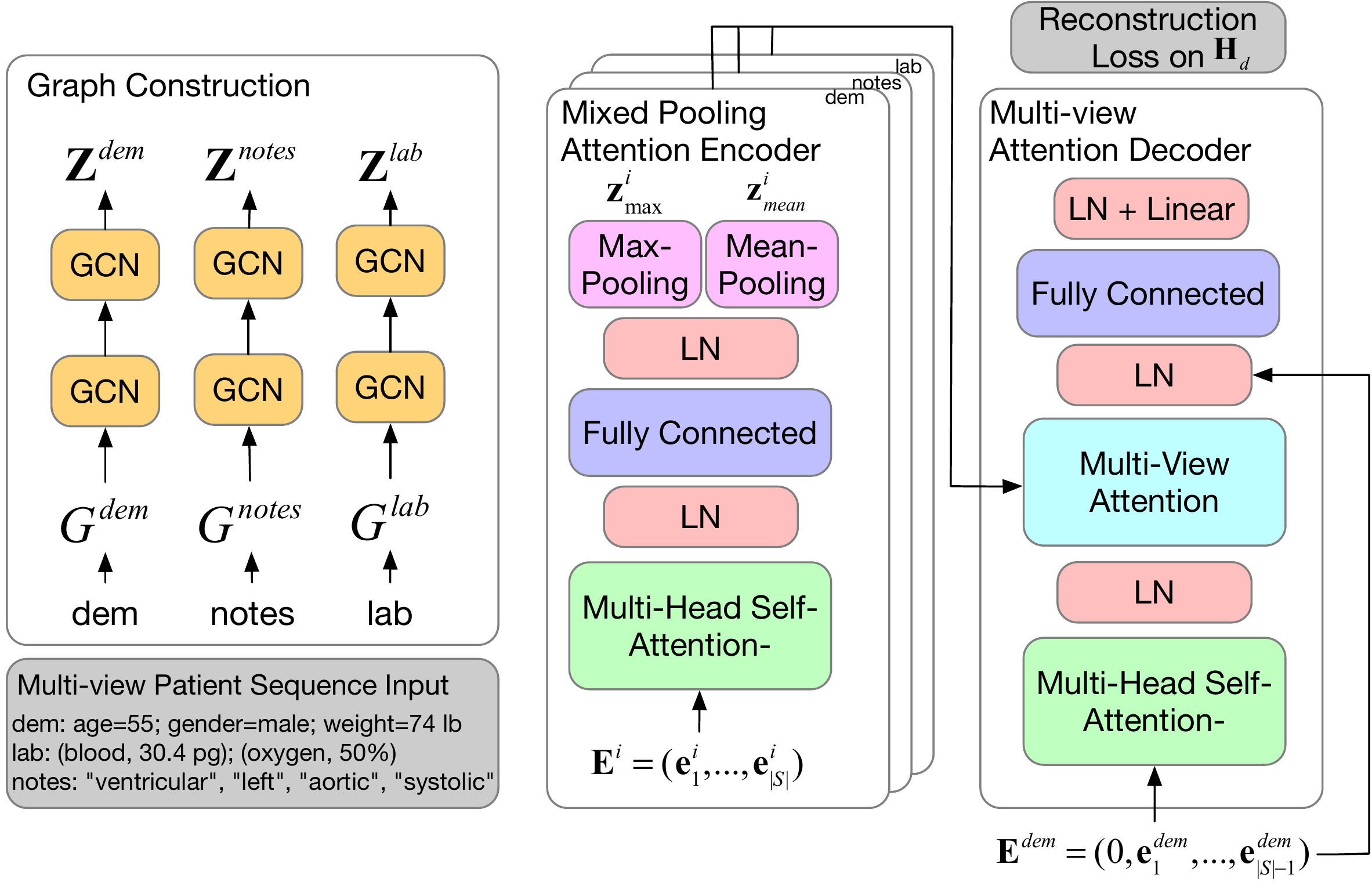}
    \caption{Proposed MPVAA Model}
    \label{fig::Figure_1}
\end{figure}

Our proposed model MPVAA, shown in Figure \ref{fig::Figure_1}, is an instance of a sequence autoencoder based on sequence to sequence learning \cite{sutskever2014sequence}. The sequence autoencoder works by first using an encoder to read the input sequence step-by-step to a hidden representation, which is then passed through a decoder to recreate the input sequence. However, unlike the traditional RNN autoencoders, MPVAA relies completely on self-attention to model input and output sequences without using RNNs or Convolution. In particular, MPVAA employs a multi-head self-attention mechanism that allows extracting  different aspects of the patient sequence into multiple vector representations. By augmenting this multi-head self-attention mechanism with a mixed pooling multi-view strategy, it further helps the model to associate heterogeneous medical information with each patient to generate a comprehensive representation specific to that patient. 

Each input patient sequence, $S$, is represented as a sequence of visits $<V_i...V_{|S|}>$, such that each visit $V_i$ is in turn a sequence of medical concepts $<m_j...m_{|V_i|}>$ occurring during that visit. Here each medical concept, $m_j \in  \mathbb{R}^{N}$, where $N$ is the total number of unique medical concepts. In order to encapsulate the patient's heterogeneous information into its learned representation, we consider three different views -- $demographic$ $information$ $(dem)$, $laboratory$ $results$ $(lab)$ and $clinical$ $notes$ $(notes)$. 
%$\{m_1,m_2,….m_{|V_i|}\}$
%$\{m_{i1},m_{i2},….m_{i{|V_i|}}\}$
In the following sections, we first discuss how the inner view embeddings are obtained, and then elaborate on the MPVAA architecture to generate patient representations. 

\subsection{Inner-View Embeddings}
For each patient, we use graph autoencoder to get the relevant embeddings of the medical concepts in each view. 

\subsubsection{Preliminary: Graph Autoencoder}
Use of graph-based neural network models like graph convolutional network (GCN) \cite{kipf2016semi} and graph auto-encoder (GAE) \cite{kipf2016variational} is becoming prevalent to learn robust representations for various applications ranging from social network analysis, bioinformatics to computer vision \cite{zhang2018graph}. A GCN learns the representations of the nodes of a graph with total \textit{N} nodes to generate output matrix $ \mathbf{Z}\in \mathbb{R}^{N \times F}$, where \textit{F} is the number of output features or latent dimensions of each node. A feature matrix, $ \mathbf{X}\in \mathbb{R}^{N \times M}$, with \textit{M} input features per node and adjacency matrix, $\mathbf{A}\in \mathbb{R}^{N \times N}$,
  are fed as inputs into the GCN to include both characteristics-based and structure-based node information. Each layer of a GCN can be summarized as:
 \begin{equation}
 \mathbf{H}_{l+1} = f(\mathbf{H}_l, \mathbf{A}, \mathbf{W}_l),
 \end{equation}
 
 where \textit{f} is a non-linear activation function and $\mathbf{H}_0 = \mathbf{X}$ and $\mathbf{H}_L = \mathbf{Z}$, for a total number of \textit{L} layers. The function, \textit{f}, and weight matrices, \textit{$\mathbf{W}_l$}, aggregate locality information to learn representation of each node. Graph auto-encoder (GAE) is an unsupervised extension of graph convolutional network which uses a GCN encoder and an inner product decoder. 

\subsubsection{Inner-View Graph Embeddings Using GAE}
A graph is constructed for each type of data in EHR from among --- \textit{demographic information} (\textit{dem}), \textit{laboratory results} (\textit{lab}) and \textit{clinical notes} (\textit{notes}) --- with each considered as a separate view. Each medical concept is a node in this inner-view network, with the similarity between two nodes forming an edge. We consider medical concepts from three different categories (i.e., disease, medication and procedure), which are extracted from the structured clinical codes. 
%$i\in \{1,\textellipsis {m}\}$

Formally, for view $i\in \{dem, lab, notes\}$, the respective graph for that view is $G^i\in {(V, \mathbf{A}, \mathbf{X}^i)}$ with total $V$ nodes, adjacency matrix $\mathbf{A}\in \mathbb{R}^{N \times N}$ and feature matrix $\mathbf{X}^i\in \mathbb{R}^{N \times M}$. $N$ is the total number of unique medical concepts and we set $M$ = $N$, so that each node feature with respect to a view is defined as the similarity relationship between two medical concepts, $m_A$ and $m_B$, in that view. The similarity is computed with Dice Coefficient as,
 \begin{equation}
\operatorname{DSC}(m_A, m_B)  = \frac{2|x^i_{m_A} 
\cap \  x^i_{m_B}|}{|x^i_{m_A}| + |x^i_{m_B}|},
\end{equation}

where \textit{$x^i_{m_A}$} and \textit{$x^i_{m_B}$} are the raw feature vectors for the respective medical concepts in that view. The more similar two medical concepts are in terms of common features in that view, the closer to 1 the dice coefficient between them will be. 

The graph auto-encoder for each view \textit{i} employs a two-layer GCN, where propagation rule is applied to get the output feature matrix, $\mathbf{Z}^i\in \mathbb{R}^{N \times {d_k}}$. It is defined as (to avoid clutter, we describe for one view as $\mathbf{Z}$):

\begin{equation}
\begin{split}
\mathbf{Z} =& f(\mathbf{X}, \mathbf{A}; \mathbf{W}_0, \mathbf{W}_1)  \\
=& softmax\left(\dot{\mathbf{A}} ReLU(\dot{\mathbf{A}}\mathbf{X}\mathbf{W}_0)\mathbf{W}_1\right).
\end{split}
\end{equation}

Here $\dot{\mathbf{A}}$ = $\hat{\mathbf{D}}^{-1/2} \hat{\mathbf{A}}\hat{\mathbf{D}}^{-1/2}$ is used for normalization, where $\hat{\mathbf{A}} = \mathbf{A} + \mathbf{I}$ is the adjacency matrix with added self-connections, $\mathbf{I}$ is the identity matrix and $\hat{\mathbf{D}}$ is the diagonal degree matrix of $\hat{\mathbf{A}}$. $\mathbf{W}_0$ and $\mathbf{W}_1$ are trainable weight matrices of first and second layers respectively. $\mathbf{A}$ in our case is a binary co-occurrence matrix of dimensions $N \times N$, such that entries with a ``1'' indicate that two medical concepts appeared within the same visit of the patient.

\subsubsection{Graph Construction Illustration for each View}
To embed the heterogeneous hospital encounter information specific to a patient, inner-view graphs are constructed for each patient. \textit{N} unique medical concepts, comprised of diagnosis, procedure and drug codes, are first extracted from the structured clinical events in the patient's EHR records. These medical concepts are modeled as the nodes in the inner-view specific graphs, $G^{dem}\in {(V, \mathbf{A}, \mathbf{X}^{dem})}$, $G^{lab}\in {(V, \mathbf{A}, \mathbf{X}^{lab})}$, and $G^{notes}\in {(V, \mathbf{A}, \mathbf{X}^{notes})}$, where \textit{V} = \textit{N}. 

%$age \in \{old, adult, neonate, middle\}$, $weight \in \{healthy, overweight, underweight\}$ and $gender \in \{male$ $, female\}$. 
We describe the graph  construction process for each view as:

\textbf{\textit{dem}}: To get feature vector $x{_m^{dem}}  \in \mathbb{R}^N$ for each medical concept $m \in N$ with respect to the view $dem$, age, weight and gender features of the patient are considered. The values for each feature are discretized into the following categorical bins, age $\in \{neonate, middle, adult, old\}$, gender $\in \{male, female\}$ and weight $\in \{healthy, overweight, underweight\}$. For the occurrence of \textit{m} in any visit of the patient, entry for the corresponding demographic features found in the patient's visit record are set to 1 in the intermediate feature vector $x_{m^{dem}}' \in \{0,1\}^9$.  $x_{m^{dem}}'$ corresponds to the features \{\textit{old}, \textit{adult}, \textit{neonate}, \textit{middle},  \textit{healthy} , \textit{overweight},   \textit{underweight},  \textit{male},  \textit{female}\}. Then \textit{DSC}(\textit{m}, $\bar{m}$), where $\bar{m} \in N$, between intermediate feature vectors $x_{m^{dem}}'$ and $x_{{\bar{m}}^{dem}}'$ are computed to fill the corresponding entry in the  feature vector $x_{m^{dem}}$.  

\textbf{\textit{lab}}: To get feature vector $x{_m^{lab}} \in  \mathbb{R}^N$ for each medical concept $\textit{m} \in \textit{N}$ with respect to the view \textit{lab}, $($\textit{lab item}, \textit{value}$)$ pair tuple features are considered. Similar to view \textit{dem}, it is checked if the concept \textit{m} occurred in any visit of the patient and entry for the corresponding laboratory results features found in the patient's visit are set to 1 in the intermediate feature vector $x_{m^{lab}}' \in \{0,1\}^g$, where \textit{g} corresponds to the total number of $($\textit{lab item}, \textit{value}$)$ tuple pairs. Then \textit{DSC}(\textit{m}, $\bar{m}$), where $\bar{m} \in \textit{N}$, between intermediate feature vectors $x_{m^{lab}}'$ and $x_{{\bar{m}}^{lab}}'$ are computed to fill the corresponding entry in the  feature vector $x_{m^{lab}}$. 

\textbf{\textit{notes}}: To get feature vector $x_{m^{notes}} \in  \mathbb{R}^N$ for each medical concept $m \in N$ with respect to \textit{notes}, UMLS\footnote{https://www.nlm.nih.gov/research/umls/} Concept Unique identifiers (CUIs) of contextual words of \textit{m} within a window, \textit{w},  in the notes are considered as the features. Notes of the patient with occurrence of \textit{m} in any visit are first extracted and entries for the CUIs of contextual words of \textit{m} appearing within \textit{w} in the notes are set to 1 in the intermediate feature vector $x_{m^{notes}}' \in \{0,1\}^h$, where \textit{h} corresponds to the vocabulary of the CUIs for the words in the clinical notes. Then DSC(\textit{m}, $\bar{m}$), where $\bar{m} \in N$, between intermediate feature vectors $x_{m^{notes}}'$ and $x_{{\bar{m}}^{notes}}'$ are computed to fill the corresponding entry in the  feature vector $x_{m^{notes}}$. 

%\begin{figure*}
%    \centering
%    \includegraphics[width=0.8\linewidth]
%    {figure.pdf}
%    \caption{The proposed HierSemCor model.}
%    \label{fig::figure_1}
%\end{figure*}

\subsection{Architecture}

To fuse the inner-view embeddings, each medical concept $m_j$ in the patient sequence $S$ is first embedded to a $d_k$ dimensional vector $e{_j^i}  = {Z}^i[m_j]$, where ${\mathbf{Z}^i} \in \mathbb{R}^{N \times {d_k}}$ is the medical concept feature matrix learned by the graph autoencoder for view $i$ and $\mathbf{Z}^i[v]$ is the $v$-th row of $\mathbf{Z}^i$. Unlike \cite{vaswani2017attention, zhang2018learning}, however, we don't add positional embeddings to the input patient embedding (i.e., $\mathbf{E}^i\in \mathbb{R}^{T \times {d_k}}$, where $T$ = $|S|\times|V_i|$) as the medical concepts within a visit form an unordered set.

MPVAA has an encoder-decoder framework and captures the cross-modal features by exploiting three types of attention: Encoder MultiHead Self-Attention, Encoder-Decoder Multi-View Attention and Decoder MultiHead Self-Attention. With the Encoder/Decoder MultiHead Self-Attention, the internal structure of the patient representation with respect to $dem$ view is captured by learning the dependencies of the medical concepts within its visits. As the $dem$ view includes the general features of a patient and is relatively more static than the other two views, we feed $\mathbf{E}^{dem}$ as the inputs into the encoder and decoder. While Encoder-Decoder Multi-View Attention facilitates the interactions among the $dem$, $lab$ and $notes$ views to generate
patient representation from a comprehensive view.

\subsubsection{Preliminary: Multi-Head Self-Attention}
The attention mechanism intends to map a query and a set of key-value pairs to an output \cite{vaswani2017attention}. The output is computed as a weighted sum of the values, where the weight assigned to each value is computed based on the query and the corresponding key. For self-attention, we use the Scaled Dot-Product Attention \cite{vaswani2017attention}:
\begin{equation}
Attention(\mathbf{Q},\mathbf{K},\mathbf{V})  = Softmax(\frac{\mathbf{Q}\mathbf{K}}{\sqrt{n}})\mathbf{V}.
\end{equation}

With MultiHead Attention, this attention mechanism is run multiple times in parallel. It is defined as:
\begin{equation}
 MultiHead(\mathbf{Q}, \mathbf{K}, \mathbf{V}) = [{head}_1;...;{head}_h]\mathbf{W}^0,
 \end{equation}
where,  
\begin{equation}
{head}_h  = Attention(\mathbf{Q}{\mathbf{W}_h^Q},\mathbf{K}{\mathbf{W}_h^K},\mathbf{V}{\mathbf{W}_h^V}).
\end{equation}
Here ${\mathbf{W}_h^Q}$, ${\mathbf{W}_h^K}$, ${\mathbf{W}_h^V}$ and ${\mathbf{W}^0}$ are parameter matrices to be learned.

\subsubsection{Mixed Pooling Attention Encoder} The encoder, shown in Figure \ref{fig::Figure_1}, converts the patient sequence embedding $\mathbf{E}^{dem} = (e_1^{dem}, e_2^{dem},…,e_{T}^{dem})$ into a hidden representation of vectors, $\mathbf{H}_e^{dem} = (h_{e_1}^{dem}, h_{e_2}^{dem},..,h_{e_{T}}^{dem})$ with two sub-layers. The first layer utilizes MultiHead Self-Attention mechanism to attend to $\mathbf{E}^{dem}$ jointly from different positions. The multiple hops of self-attention enables it to learn multiple vector representations of the patient focusing on different parts of the patient's visit sequence $\mathbf{E}^{dem}$. For example, each part can be a component capturing the related medical concepts within the visits, which reflects a semantic aspect of the patient's hospital profile. Thus the overall semantics can be represented by the multiple vector representations computed by the multi-head self-attention. It is defined as

\begin{equation}
\begin{split}
\mathbf{A}_{e^{dem'}} &= MultiHead(\mathbf{E}^{dem}\mathbf{W}_q^{dem},\\
&\mathbf{E}^{dem}\mathbf{W}_k^{dem},\mathbf{E}^{dem}\mathbf{W}_v^{dem}),
\end{split}
\end{equation}

where ${\mathbf{W}_q^{dem}} \in  \mathbb{R}^{{d_m} \times d_k}$, ${\mathbf{W}_k^{dem}} \in  \mathbb{R}^{{d_m} \times d_k}$ and ${\mathbf{W}_v^{dem}} \in  \mathbb{R}^{{d_m} \times d_v}$ are parameter matrices;

The second layer, on the other hand, is a fully connected feed-forward network that applies a non-linear activation (i.e., ReLU) on the linearly transformed outputs from the first layer, followed by another linear transformation. We apply residual connections after each layer. So the hidden representation $\mathbf{H}_e^{dem}$ is produced with the following equations:
\begin{equation}
\mathbf{A}_e^{dem} = LayerNorm(\mathbf{A}_e^{dem'} + \mathbf{E}^{dem}),
\end{equation}

\begin{equation}
\mathbf{H}_{e}^{dem'} = ReLU(\mathbf{A}_{e}^{dem}\mathbf{W}_{\mathbf{E}_a}^{dem} + \mathbf{b}_{e_a}^{dem})\mathbf{W}_{\mathbf{e}_b}^{dem} +
\mathbf{b}_{e_b}^{dem},
\end{equation}

\begin{equation}
\mathbf{H}_{e}^{dem} = LayerNorm(\mathbf{H}_e^{dem'} + \mathbf{A}_e^{dem}),
\end{equation}

where ${\mathbf{W}_{e_a}^{dem}} \in  \mathbb{R}^{{d_m} \times d_f}$ and ${\mathbf{W}_{e_b}^{dem}} \in  \mathbb{R}^{{d_f} \times d_m}$ are parameter matrices; ${\mathbf{b}_{e_a}^{dem}}\in  \mathbb{R}^{d_f}$ and ${\mathbf{b}_{e_b}^{dem}}\in  \mathbb{R}^{d_m}$ are bias vectors;
LayerNorm denotes layer normalization and ReLU is activation function.

To aggregate the hidden representations $\mathbf{H}_e^{dem}$ to a fixed dimensional vector, a mixed pooling strategy is performed that produces the mixed representation $\mathbf{z}^{dem}$. It does so by a randomly weighting technique that measures the importance of the mean representation and the max representation. This generalized pooling approach results in enriching the expressiveness of the attention mechanism. 

In max pooling, max operation is applied on each vector of the hidden representation to extract the most salient feature pertaining to that time step. While mean pooling takes into consideration all the features and summarizes them into a global representation:    
\begin{equation}
\mathbf{z}_{max} = [max(\mathbf{h}^{dem}_{e_1}),max(\mathbf{h}^{dem}_{e_2}),....,max(\mathbf{h}^{dem}_{e_{T}})],
\end{equation}
%\begin{equation}
%${z_{max}} =
%\max\limits_{\forall t \in {|S|}} h^i_{e_t}$
%\end{equation}
%\begin{equation}
%${z_{max}} =
%\forall t \in {|S|} max( h^i_{e_t})$
%\end{equation}
\begin{equation}
{\mathbf{z}_{mean}} = \frac{1}{T}\sum\limits_{t}{\mathbf{h}^{dem}_{e_t}}.
\end{equation}

The mixed pooling strategy is defined as
\begin{equation}
\mathbf{z}^{dem} = \lambda\mathbf{z}_{max} + (1-\lambda)\mathbf{z}_{mean},
\end{equation}
where $\lambda$ is a random value between 0 and 1 indicating the weighted contribution of the mean pooling and max pooling methods in the final representation. 

From patient perspective, the mixed pooling operation allows to simultaneously encode the most activated dimension in the embedding space of each medical concept occurring in their visit sequence, and also provide a comprehensive capture of semantics across all the medical concept vectors. This can greatly ease interpretability of the learned embeddings as well.

\subsubsection{Multi-View Attention Decoder for Patient Representation}

 As we intend to learn general-purpose patient representations with our model that should exhibit comparable performance across all downstream tasks in healthcare, it is important to integrate a holistic view of heterogeneous data associated with each patient. Henceforth, given the mixed patient representation, $\mathbf{z}^{dem}$ for view $dem$, the purpose of the decoder is to reconstruct the input sequence so as to incorporate the relevant attributes from $lab$ and $notes$ views as well. By connecting the encoder and decoder with a multi-view attention module, it is possible to fuse together the embedddings from different views into the hidden representation of the patient sequence.

As shown in Figure \ref{fig::Figure_1}, first the input patient embedding, $\mathbf{E}^{dem} = (e^{dem}_1, e^{dem}_2,..,e^{dem}_{T})$, is shifted right as $\mathbf{E}^{dem'} = (0, e_1^{dem}, . . . , e_{T-1}^{dem})$ to get the decoder input. Equations (7) and (8) are then employed in the decoder MultiHead Self-Attention layer to get $\mathbf{A}_{d_{sa}} = (a_{d_{{sa}_1}},...,a_{d_{{sa}_T}})$.

Following the self-attention layer, multi-view attention is computed using the cross-view representation $\mathbf{C}$ as,

\begin{equation}
{\mathbf{A}_d'} = MultiHead(\mathbf{A}_{d_{sa}},\mathbf{C},\mathbf{C}),
\end{equation}

\begin{equation}
{\mathbf{A}_d} = LayerNorm({\mathbf{A}_d'} + \mathbf{A}_{d_{sa}}),
\end{equation}

\begin{equation}
{\mathbf{H}_d'} = ReLU(\mathbf{A}_d{\mathbf{W}_{d_a}} + {\mathbf{b}_{d_a}}){\mathbf{W}_{d_b}} + {\mathbf{b}_{d_b}},
\end{equation}

\begin{equation}
{\mathbf{H}_d} = LayerNorm(\mathbf{H}_d' + \mathbf{A}_d),
\end{equation}
where ${\mathbf{W}_{d_a}} \in  \mathbb{R}^{{d_m} \times d_f}$ and ${\mathbf{W}_{d_b}} \in  \mathbb{R}^{{d_f} \times d_m}$ are parameter matrices; ${\mathbf{b}_{d_a}}\in  \mathbb{R}^{d_f}$ and ${\mathbf{b}_{d_b}}\in  \mathbb{R}^{d_m}$ are bias vectors.

The cross-view representation $\mathbf{C}$ incorporates the interactions among all the views. To calculate it, first the weighted mixed representations of the other two views, $\mathbf{z}^{lab}$ and $\mathbf{z}^{notes}$, are applied to $\mathbf{z}^{dem}$ with a non-linear activation. Mathematically it is denoted as,

\begin{equation}
\mathbf{C'} = f({\mathbf{z}^{dem}}\cdot {\mathbf{W}_{z}^{dem}} + {\mathbf{z}^{lab}}\cdot {\mathbf{W}_{z}^{lab}} + {\mathbf{z}^{notes}}\cdot {\mathbf{W}_{z}^{notes}}),
\end{equation}

where $\mathbf{W}_{z}^{dem} \in  \mathbb{R}^{{d_m} \times d_m}$, $\mathbf{W}_{z}^{lab} \in  \mathbb{R}^{{d_m} \times d_m}$ and  $\mathbf{W}_{z}^{notes} \in  \mathbb{R}^{{d_m} \times d_m}$ are parameter matrices; $f$ is the non-linear activation function (i.e., tanh). 

$\mathbf{C}'$ is then
fed into a softmax layer and finally the softmax output is combined with the hidden representation $\mathbf{A}_{d_{sa}}$  via element-wise multiplication $\otimes$:
\begin{equation}
\mathbf{C} = Softmax(\mathbf{C}') \otimes {\mathbf{A}_{d_{sa}}}.
\end{equation}

Essentially, the generation of $\mathbf{C}$ through this weighted mean, where the weight indicates the relevance of the view with regards to the patient, leads to capturing the contributions of the heterogeneous data types into the final patient representation.

The probability of
generating the whole patient sequence is then calculated as:
\begin{equation}
P(m_j|m_1,....,{m_{j-1}}, {\mathbf{h}_{d_j}}) \propto exp(\mathbf{W}_p{\mathbf{h}_{d_j}} + b_p).
\end{equation}

The objective is the sum of the log-probabilities for the
input sequence itself:

\begin{equation}
J(\theta) = \sum\limits_{j}
logP(m_j|m_1, . . . , {m_{j-1}}, {\mathbf{h}_{d_j}}).
\end{equation}

MPVAA learns to reconstruct the input
sequence by optimizing the objective in equation 21.

\section{Experiments}

\subsection{Settings}
We evaluate our model on the publicly available MIMIC-III dataset \cite{johnson2016mimic}. This database contains de-identified clinical records for $>$ 40K patients admitted to critical care units over 11 years. It contains a wide range of heterogeneous healthcare information such as demographics, laboratory test results, procedures, medications, diagnosis codes, nurse and physician notes, among others. ICD-9 codes for diagnosis/procedures and NDC codes for medications were extracted from patients with at least two visits to construct each graph. Tables \ref{tab::Table5} and \ref{tab::Table6} outline other statistics about the data.

\begin{table}[]
\centering
\caption{Data Statistics Summary}
\label{tab::Table5}
\begin{tabular}{ll}
\toprule
\textbf{MIMIC-III} & \textbf{VALUE}\\ \midrule
\# of patients                 & 7,499  \\
\# of visits                & 19,911  \\ 
avg. \# of visits per patient     & 2.66\\ 
\# of unique ICD9 codes  & 4,893 \\
avg. \# of codes per visit  & 13.1 \\
max \# of codes per visit  & 39 \\
\bottomrule
\end{tabular}
\end{table}

\begin{table}[]
\centering
\caption{Summary of Clinical Codes used}
\label{tab::Table6}
\begin{tabular}{llll}
\toprule
\textbf{MIMIC-III} & \textbf{VALUE}\\ \midrule
\# unique clinical codes                & 11135  \\
\# of unique ICD-9 diagnoses codes                & 4894  \\ 
\# of unique ICD-9 procedure codes    & 2032\\ 
\# of unique NDC medication codes  & 4209 \\
\bottomrule
\end{tabular}
\end{table}

% \begin{table}[]
% \centering
% \caption{Summary of HF Dataset}
% \label{tab::Table7}
% \begin{tabular}{llll}
% \toprule
%                      & 	MIMIC-III\\ \midrule
% total data size               & 1485  \\
% \# train                & 1113  \\ 
% \# test    & 186\\ 
% \# valid  & 186 \\
% \bottomrule
% \end{tabular}
% \end{table}

\subsection{Implementation and Training Details}
For each medical concept, the feature matrix for $dem$ and $lab$ views in graph auto-encoder is constructed by considering general information about the patient such as age, weight, gender from ADMISSIONS table and laboratory measurements results from LABEVENTS table respectively in MIMIC-III, that occurred with the concept. For $notes$ view, Unified Medical Language System (UMLS) concepts, obtained with MetaMap 2, for context within a window of the medical concept mention are considered as the features. 

For the views $dem$ and $lab$, there were no instances with missing features and each medical concept node in the graph had all feature value pairs. While in the case of notes view, if the name of the
medical concept was not found in the notes, then context of other co-occurring concepts were considered as its features. We used only ``discharge summary" notes and the value of window, $w$, was
set to 2.

We used 5 parallel attention heads and set the final patient representation dimension to 500 for fair comparison against the baselines. The proposed model is trained using Adam \cite{kingma2014adam} in minibatches
of 64 with a learning rate of 0.001. $\lambda$ is shared among all the views.
\subsection{Evaluation Tasks and Metrics}
The learned patient representations are evaluated on the following two extrinsic medical tasks,

\textbf{Outcome Prediction}: This is a binary prediction task that tries to predict whether the patient is at risk of developing a disease in the future visit, $v_{t+1}$, trained on visit embedding sequence up to $v_t$. We focus on patients with heart failure (HF) disease. Thereby, we examine only patients with at least two visits and check if they contain an occurrence of heart failure in their $v_{t+1}$ visit. These are considered as the instances belonging to the positive class (HF). As average number of visits per patient is 2 in MIMIC-III, $v_{t+1}$ in our case is 1. A binary logistic regression classifier is trained/tested to perform this prediction task. 
The train/test/validation split for positive instances 
% are shown in  Table \ref{tab::Table7}, 
is (1113/186/186)
and the same split is applied to equal number of total negative instances, where negative instances are formed with patients who do not have HF code in their record up to the $v_{t+1}$ visit. 

\textbf{Sequential Disease Prediction:} The goal is to predict all the diagnosis codes of the next visit $v_{t+1}$ at every time step, having trained on visit sequence up to the $t$-th visit. It can be considered as a multi-label classification task.

We report performance on Heart Failure Prediction task assessed based on AUC-ROC, AUC-PR and Accuracy, while Sequential Disease Prediction task is evaluated in terms of Normalized Cumulative Discounted Gain (\textit{NCDG}). To evaluate the quality of the predicted diagnosis codes, they are first sorted according to their prediction values and \textit{NCDG} is measured at different cutoff values, \textit{k} = 5, 15 and 25, against the true diagnosis codes. 

\textit{NCDG} is computed as the ratio of \textit{DCG} and \textit{Ideal} \textit{DCG} (\textit{IDCG}) defined by the following equations,
\begin{equation}
NCDG = \frac{DCG(true, pred)}{IDCG},
\end{equation}
\begin{equation}
DCG(x,y) = \sum\limits_{i=1}^{|x|}{\frac{{2^{rel_i}}-1}{\log_2(i+1)}},
\end{equation}
where $rel_i$ is the rank of $i$-th element in \textit{x} relative to the sorted predictions (\textit{y}) and $IDCG = DCG(true, true)$. \textit{True} and \textit{pred} refer to the list of true and predicted diagnosis codes respectively.

\begin{figure*}[bth!]
    \begin{minipage}{.5\textwidth}
    \centering
    \includegraphics[width=\linewidth]{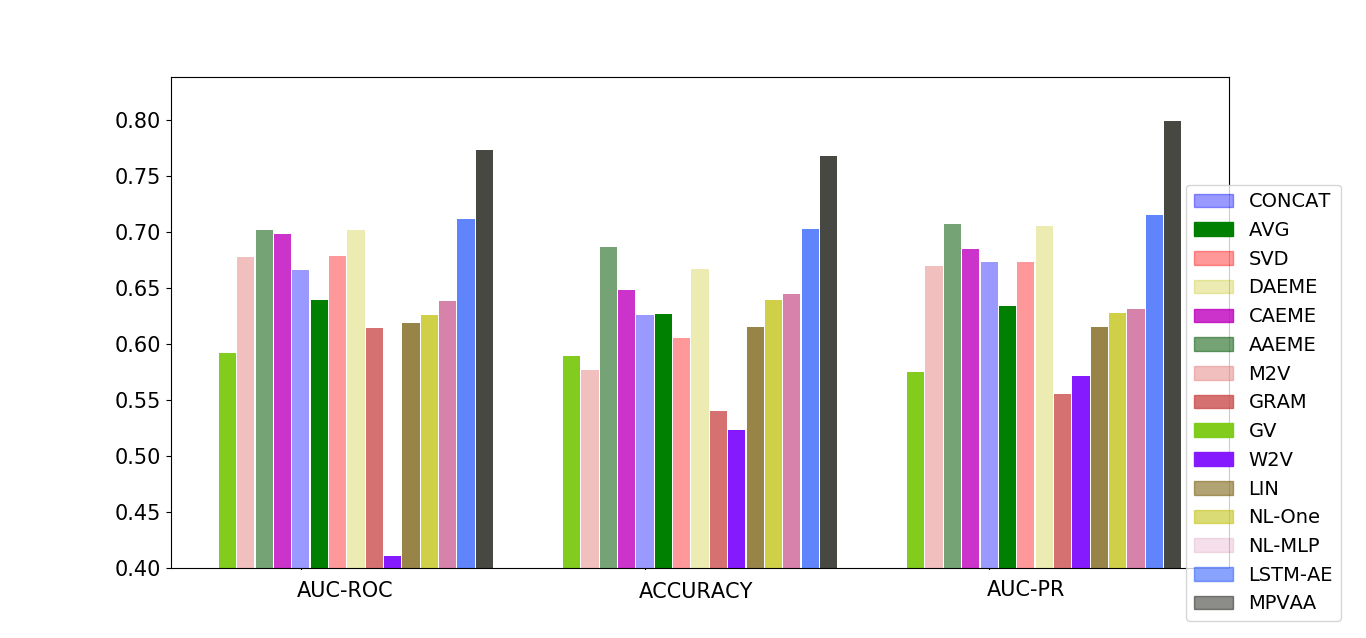}
    % \caption{Performance of different embeddings on Heart Failure \textbf{(HF)} Prediction task evaluated with metrics AUC-ROC, Accuracy and AUC-PR}
    \caption{Performance on Heart Failure (HF) Prediction task in AUC-ROC, Accuracy and AUC-PR}
    \label{fig::Figure_res_1}
    \end{minipage}
    \begin{minipage}{.5\textwidth}
    \centering
    \includegraphics[width=\linewidth]{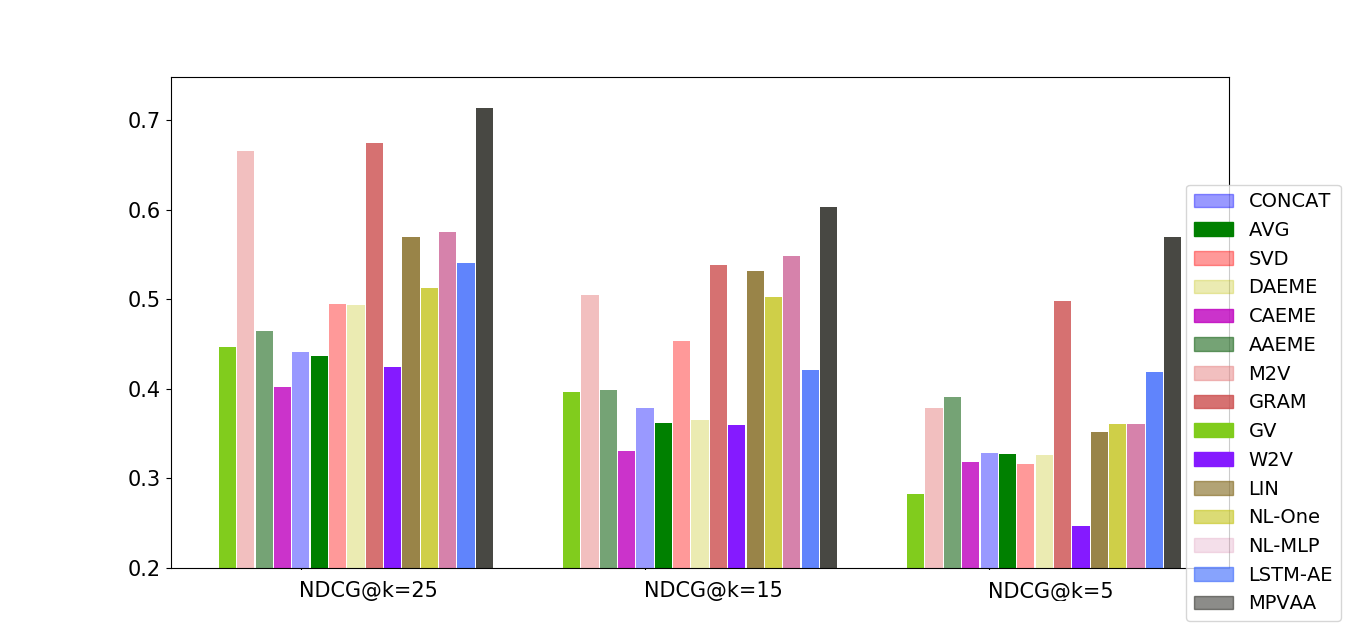}
    % \caption{Performance of different embeddings on  Sequential Disease Prediction Task evaluated with NDCG@k=25,15,5}
    \caption{Performance on Sequential Disease Prediction task in NDCG@k=25,15,5}
    \label{fig::Figure_res_2}
    \end{minipage}
\end{figure*}

\subsection{Baseline Models}
We compare MPVAA against three categories of baseline models (i.e., inside parentheses):
\textbf{1)} the aggregation method used to fuse the embeddings from the different views (CONCAT, AVG, SVD, DAEME, CAEME, AAEME)
\textbf{2)} state-of-the-art embedding models (M2V, GRAM, GV, W2V, LIN, NL-One, NL-MLP)
\textbf{3)} use of rnn units (LSTM-AE).

For the baselines, each visit $v_t$ of the patient sequence, $S$ $\{v_1,v_2,…..,v_{|S|}\}$, is represented with a list of medical concepts, $\{m_{1t},m_{2t},…..,m_{T}\}$, where they differ in how the embedding of the medical concept is learned. We consider $N$ as the total number of unique medical codes (i.e., ICD-9, NDC) present across all the patients records. For the baseline models under category 1), the learned inner-view embeddings for the medical concepts are considered as we intend to focus on the effect on performance from the embedding aggregation method.

\textbf{CONCAT}: The embeddings for the three views are simply concatenated to get the final embedding of each medical concept. We $\ell_{2}$ normalize each embedding before concatenation to ensure that each embedding contributes equally. 

\textbf{AVG}: The embeddings for the three views are averaged to represent the final embedding of each medical concept. $\ell_{2}$ normalization is similarly performed.

\noindent\textbf{SVD}: Consider a matrix $C$ of dimension $N \times k$, where \textit{k} is dimension of the resulting embedding from the concatenation of the $\ell_{2}$ normalized embeddings for each medical concept. Singular Value Decomposition (SVD) \cite{golub1971singular} is applied on $C$ to get the decomposition \textit{C} = $USV^T$. For each concept, the corresponding vector in \textit{U} is considered as the SVD embedding.

\noindent\textbf{DAEME}:  Decoupled Autoencoded Meta-Embedding \cite{bollegala2018learning} uses separate set of encoder-decoder for each view. The encoded representations are concatenated and then the individual components are reconstructed from corresponding decoders. Each autoencoder is implemented as a single layer neural network.

\noindent\textbf{CAEME}: Concatenated Autoencoded Meta-Embedding \cite{bollegala2018learning} is similar to DAEME, except that the reconstruction is done from the concatenation of the encoded representations.

\noindent\textbf{AAEME}: Averaged Autoencoded Meta-Embedding \cite{bollegala2018learning} is similar to CAEME, except that the reconstruction is done from the average of the encoded representations.

\noindent\textbf{M2V}: Med2Vec \cite{choi2016multi} is a two-layer neural network for learning lower dimensional representations for medical concepts.

\noindent\textbf{GRAM}: \cite{choi2017gram} is a graph-based attention model augmented with knowledge from medical ontology. 

\noindent\textbf{GV}: GloVe \cite{pennington2014glove} is an unsupervised learning approach of word embeddings based on word co-occurrence matrix. In our case, the embedding matrix is of dimension $N \times k$, where $k$ is the dimension of Glove embeddings. 

\noindent\textbf{W2V}: The Skip-gram model \cite{mikolov2013distributed} learns word representations based
on the co-occurrence information of words within a context window of a predefined size. In our case, the embedding matrix is of dimension $N \times k$, where $k$ is the dimension of Word2Vec embeddings. 

\noindent\textbf{LIN}: A visit $v_t$ is represented with binary, multi-hot encoding as $v_t \in {0,1}^{N}$. That is, only the dimension corresponding to the code is set to 1. This vector is linearly transformed with $W \in \mathbb{R}^{d\times {N}}$ embedding matrix with embedding dimension \textit{d} so that $v'_t = {W}{v_t}$. Embedding in \textit{W} in initialized with Glove \cite{pennington2014glove}.

\textbf{NL-One}: It adds non-linearity to visit $v'_t$ by passing it through a non-linear activation $\sigma$. We used ReLU activation function, $v'_t = \sigma ({W}{v_t})$. 

\noindent\textbf{NL-MLP}: It is the same as Non-Linear(one) but has one more layer to increase expressivity. 

\noindent\textbf{LSTM-AE}: The patient sequence representation $S$ is learned through an autoencoder for each view. The encoder and decoder are implemented with RNN units (i.e., LSTM). The final patient representation is considered as the aggregation of the learned representations from all the views.

We further compare the performance of the proposed MPVAA against ablated versions of the model to exhibit contributions from each component. 

\noindent\textbf{MMAA}: The Mean-Max
Attention Autoencoder \cite{zhang2018learning} using mean-max attention during decoding.

\noindent\textbf{MMVAA}: The proposed model doing mean-max pooling operation on the encoded representation, $H^i_e$.

\noindent\textbf{VAA}: The proposed model without doing mixed pooling operation on the encoded representation, $H^i_e$.

\noindent\textbf{MPVAA-Sin}: The proposed model with positional encodings added to input embeddings, $E^i$.

\subsection{Evaluation Results}
The experimental results for the heart failure prediction and sequential disease prediction tasks are shown in Figures \ref{fig::Figure_res_1} and \ref{fig::Figure_res_2} respectively. We do comparative analysis on the performance between the proposed and the three different categories of the baseline representation learning models (i.e., 1), 2) and 3)). A key observation is that the proposed MPVAA consistently outperforms all baselines in all three categories for both tasks across all the metrics. We first evaluate how effective the different fusion approaches are in aggregating the embeddings from the different views. Among the general ensemble methods (i.e., CONCAT, AVG and SVD), performing a global projection on the concatenated embeddings through SVD is shown to give better results across most metrics than the other two for both tasks. Now comparing the baselines implementing autoencoder (i.e., DAEME, CAEME and AAEME), it can be seen that they do better than their simple ensemble counterparts, CONCAT and AVG respectively. We can attribute this to the dimensionality reduction they enforce through the reconstruction of the hidden representation, that results in embedding the key features of the input into its learned representation and establishes autoencoders as a comparable base model for representation learning. This justifies our choice of basing MPVAA on the autoencoder framework. 

Comparing against the second category of baselines, in particular, MPVAA shows a 13\% and 3.5\% performance gain against M2V and GRAM in HF prediction and Sequential Disease Prediction tasks respectively. The good results of M2V and GRAM compared to the other baselines in this category can be attributed to the fact that they are specifically designed for learning representations from EHR. This means that training on domain-specific data (e.g., EHR) strengthens generalizability of the representations as opposed to on general data. 

MPVAA completely relies on attention mechanism to model patient representations. Its better performance on both tasks than LSTM-AE verifies this contribution compared to the use of recurrent units (i.e., LSTM) in LSTM-AE. One reason for this could be that RNNs put emphasis on information towards the end of the sequence and hence is not able to connect information from past visits.

\begin{figure*}[bth!]
    \begin{minipage}{.5\textwidth}
    \centering
    \includegraphics[width=\linewidth]{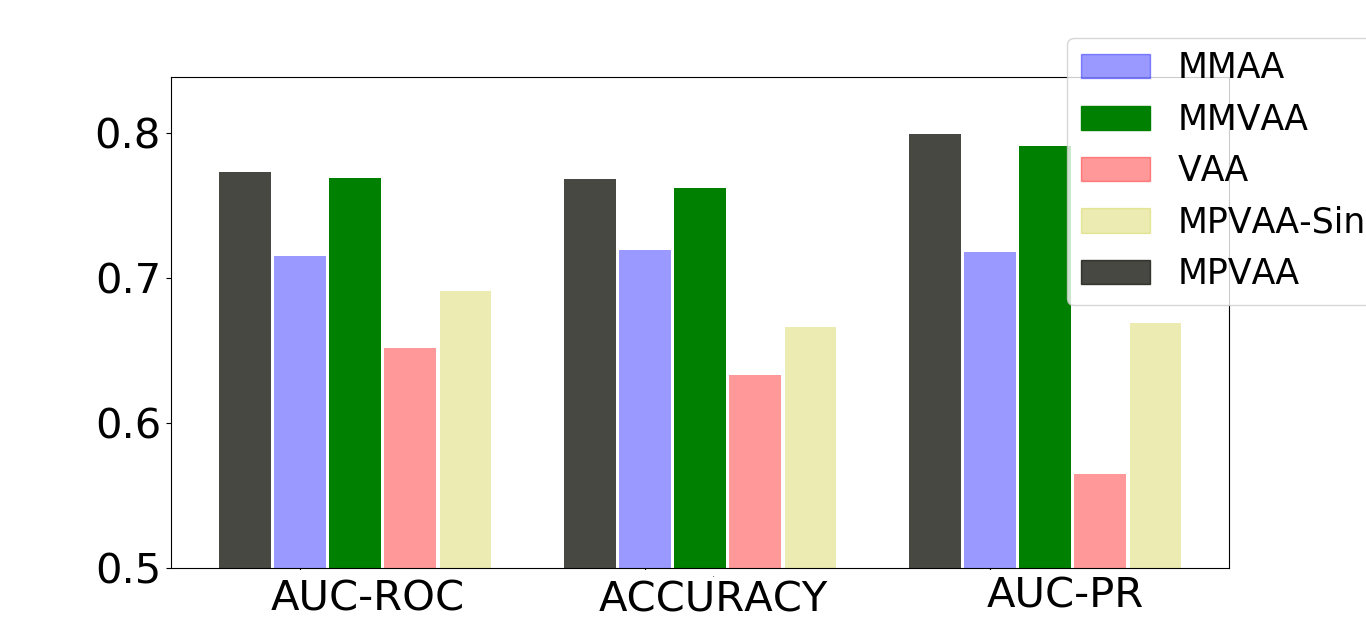}\caption{Ablation Performance of MPVAA on Heart Failure (HF) Prediction in AUC-ROC, Accuracy and AUC-PR}
    \label{fig::Figure_res_3}
    \end{minipage}
    \begin{minipage}{.5\textwidth}
    \centering
    \includegraphics[width=\linewidth]{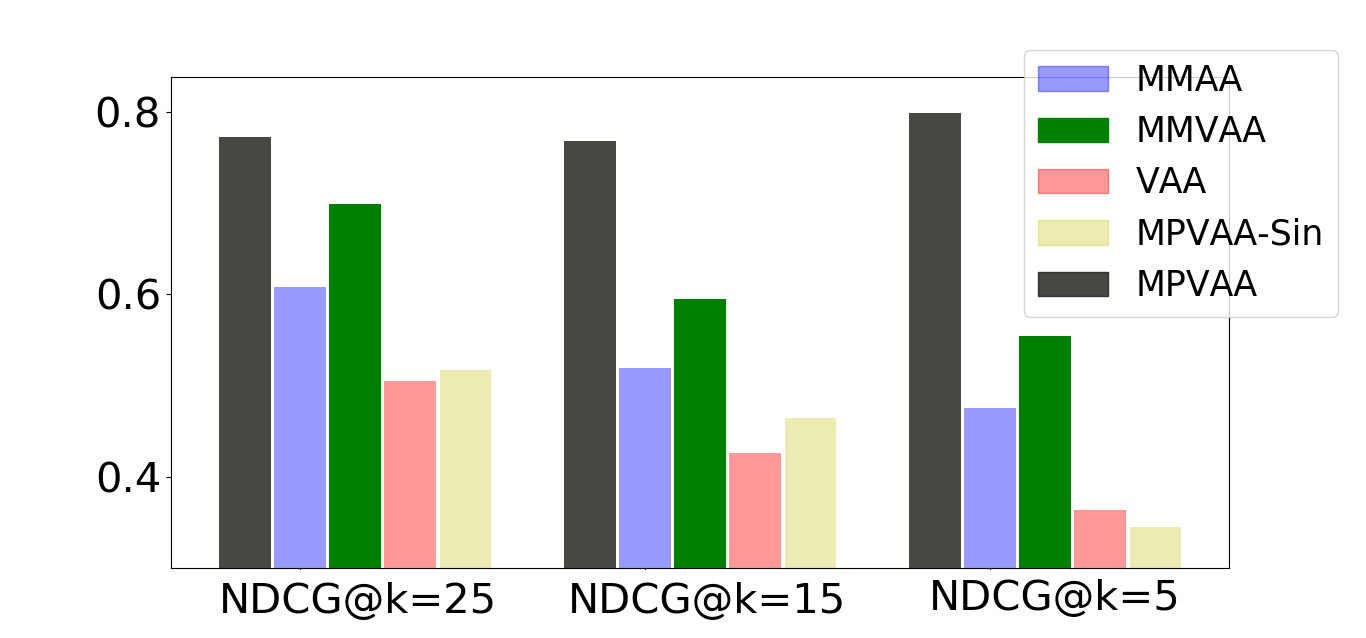}\caption{Ablation Performance of MPVAA on Sequential Disease Prediction in NDCG@k=25,15,5}
    \label{fig::Figure_res_4}
    \end{minipage}
\end{figure*}
\subsection{Ablation Study}
To quantitatively evaluate the effect of various components used in MPVAA on the model performance, we compare the performance of MPVAA against its variants (i.e., MMAA, MMVAA, VAA and MPVAA-Sin) as presented in Figures \ref{fig::Figure_res_3} and \ref{fig::Figure_res_4} respectively for Heart Failure Prediction and Sequential Disease Prediction tasks. It can be seen that all four variants gave worse performance than MPVAA for both the tasks. VAA excludes mixed pooling operation of the encoded representation and causes the most decline in performance compared to the full model, MPVAA. While MMVAA uses  deterministic mean-max pooling instead of the stochastic approach used in MPVAA and is similarly outperformed. This asserts the stochastic mixed pooling component of our proposed MPVAA as an integral part of it. We believe that since MMAA does not incorporate the interactions among the patient information from different views while decoding as opposed to the use of multi-view attention in MPVAA, it is not able to embed the dependencies among the different views in the encoded embeddings. As mentioned earlier, the medical codes within each visit are unordered, so adding positional embeddings does not improve performance with MPVAA-Sin. 

\section{Related Works}
As healthcare data is growing owing to the increase in hospital use of EHR, research using EHR has become an active area to mine useful associations between different clinical variables, that unravel information critical to fulfilling each patient's medical needs alongside facilitating different real-world clinical predictive tasks. The cornerstone of effective implementation of EHR lies in robust representation learning that should holistically capture semantic relations among heterogeneous medical entities. Use of raw EHR data in terms of hand-engineered feature statistics (e.g., count) or binary representation with hot vector as inputs to predictive models is labor intensive for the former and fail to capture the hierarchical latent relationships at all levels, paving the way for learning distributed vectors through deep learning recently. A regularized nonnegative Restricted Boltzmann Machine is formulated in \cite{tran2015learning} that embeds the medical events in EHR to low-dimensional space, while a stack of denoising autoencoders (AE) and multi-layer perceptron (MLP) are used in \cite{miotto2016deep} and \cite{choi2016multi} respectively to get patient vectors. More common has been the use of Recurrent Neural Networks (RNN) to capture the sequential nature of EHR records for different predictive tasks \cite{choi2016doctor,esteban2016predicting}. \cite{choi2016retain} adds a two-level attention mechanism to RNN to get more interpretable representation. A graph-based attention model combined with RNN is used in \cite{choi2017gram} to address data insufficiency issue by augmenting knowledge from medical ontology. However, unlike all the aforementioned works, MPVAA first learns graph-based inner-view embeddings from the different types of data in EHR, and then captures their cross-modal interactions with multi-view attention into a holistic representation, which has shown superior performance.

\section{Conclusion}
In this paper, we present an unsupervised framework, Mixed Pooling Multi-View Attention Autoencoder (MPVAA), for learning patient representations. To enforce personalization in the learning that is specific to each patient, inner-view graphs are constructed. It facilitates better modeling by exploiting the complementary information from multiple data modalities in EHR. A combination of powerful attention mechanisms (i.e., self-attention, multi-view attention) is employed to capture the interactions among the cross-modal features. Comprehensive experiments performed demonstrate that the proposed MPVAA model outperforms the state-of-the-art baselines on Heart Failure Prediction and Sequential Disease Prediction tasks.

%To enforce personalization in the learning that is specific to each patient, multi-view graphs are constructed from each patient's heterogeneous medical records.

%develop a robust system for representation learning in healthcare

\bibliography{thebibliography}

\begin{thebibliography}{10}

\bibitem{bollegala2018learning}
D.~Bollegala and C.~Bao.
\newblock Learning word meta-embeddings by autoencoding.
\newblock In {\em Proceedings of the 27th International Conference on
  Computational Linguistics}, pages 1650--1661, 2018.

\bibitem{choi2016doctor}
E.~Choi, M.~T. Bahadori, A.~Schuetz, W.~F. Stewart, and J.~Sun.
\newblock Doctor ai: Predicting clinical events via recurrent neural networks.
\newblock In {\em Machine Learning for Healthcare Conference}, pages 301--318,
  2016.

\bibitem{choi2016multi}
E.~Choi, M.~T. Bahadori, E.~Searles, C.~Coffey, M.~Thompson, J.~Bost,
  J.~Tejedor-Sojo, and J.~Sun.
\newblock Multi-layer representation learning for medical concepts.
\newblock In {\em Proceedings of the 22nd ACM SIGKDD International Conference
  on Knowledge Discovery and Data Mining}, pages 1495--1504. ACM, 2016.

\bibitem{choi2017gram}
E.~Choi, M.~T. Bahadori, L.~Song, W.~F. Stewart, and J.~Sun.
\newblock Gram: graph-based attention model for healthcare representation
  learning.
\newblock In {\em Proceedings of the 23rd ACM SIGKDD International Conference
  on Knowledge Discovery and Data Mining}, pages 787--795. ACM, 2017.

\bibitem{choi2016retain}
E.~Choi, M.~T. Bahadori, J.~Sun, J.~Kulas, A.~Schuetz, and W.~Stewart.
\newblock Retain: An interpretable predictive model for healthcare using
  reverse time attention mechanism.
\newblock In {\em Advances in Neural Information Processing Systems}, pages
  3504--3512, 2016.

\bibitem{choi2016medical}
E.~Choi, A.~Schuetz, W.~F. Stewart, and J.~Sun.
\newblock Medical concept representation learning from electronic health
  records and its application on heart failure prediction.
\newblock {\em arXiv preprint arXiv:1602.03686}, 2016.

\bibitem{esteban2016predicting}
C.~Esteban, O.~Staeck, S.~Baier, Y.~Yang, and V.~Tresp.
\newblock Predicting clinical events by combining static and dynamic
  information using recurrent neural networks.
\newblock In {\em Healthcare Informatics (ICHI), 2016 IEEE International
  Conference on}, pages 93--101. Ieee, 2016.

\bibitem{frome2013devise}
A.~Frome, G.~S. Corrado, J.~Shlens, S.~Bengio, J.~Dean, T.~Mikolov, et~al.
\newblock Devise: A deep visual-semantic embedding model.
\newblock In {\em Advances in neural information processing systems}, pages
  2121--2129, 2013.

\bibitem{golub1971singular}
G.~H. Golub and C.~Reinsch.
\newblock Singular value decomposition and least squares solutions.
\newblock In {\em Linear Algebra}, pages 134--151. Springer, 1971.

\bibitem{johnson2016mimic}
A.~E. Johnson, T.~J. Pollard, L.~Shen, H.~L. Li-wei, M.~Feng, M.~Ghassemi,
  B.~Moody, P.~Szolovits, L.~A. Celi, and R.~G. Mark.
\newblock Mimic-iii, a freely accessible critical care database.
\newblock {\em Scientific data}, 3:160035, 2016.

\bibitem{kingma2014adam}
D.~P. Kingma and J.~Ba.
\newblock Adam: A method for stochastic optimization.
\newblock {\em arXiv preprint arXiv:1412.6980}, 2014.

\bibitem{kipf2016semi}
T.~N. Kipf and M.~Welling.
\newblock Semi-supervised classification with graph convolutional networks.
\newblock {\em arXiv preprint arXiv:1609.02907}, 2016.

\bibitem{kipf2016variational}
T.~N. Kipf and M.~Welling.
\newblock Variational graph auto-encoders.
\newblock {\em arXiv preprint arXiv:1611.07308}, 2016.

\bibitem{mikolov2013distributed}
T.~Mikolov, I.~Sutskever, K.~Chen, G.~S. Corrado, and J.~Dean.
\newblock Distributed representations of words and phrases and their
  compositionality.
\newblock In {\em Advances in neural information processing systems}, pages
  3111--3119, 2013.

\bibitem{miotto2016deep}
R.~Miotto, L.~Li, B.~A. Kidd, and J.~T. Dudley.
\newblock Deep patient: an unsupervised representation to predict the future of
  patients from the electronic health records.
\newblock {\em Scientific reports}, 6:26094, 2016.

\bibitem{pennington2014glove}
J.~Pennington, R.~Socher, and C.~Manning.
\newblock Glove: Global vectors for word representation.
\newblock In {\em Proceedings of the 2014 conference on empirical methods in
  natural language processing (EMNLP)}, pages 1532--1543, 2014.

\bibitem{shickel2017deep}
B.~Shickel, P.~J. Tighe, A.~Bihorac, and P.~Rashidi.
\newblock Deep ehr: a survey of recent advances in deep learning techniques for
  electronic health record (ehr) analysis.
\newblock {\em IEEE journal of biomedical and health informatics},
  22(5):1589--1604, 2017.

\bibitem{sutskever2014sequence}
I.~Sutskever, O.~Vinyals, and Q.~V. Le.
\newblock Sequence to sequence learning with neural networks.
\newblock In {\em Advances in neural information processing systems}, pages
  3104--3112, 2014.

\bibitem{tran2015learning}
T.~Tran, T.~D. Nguyen, D.~Phung, and S.~Venkatesh.
\newblock Learning vector representation of medical objects via emr-driven
  nonnegative restricted boltzmann machines (enrbm).
\newblock {\em Journal of biomedical informatics}, 54:96--105, 2015.

\bibitem{vaswani2017attention}
A.~Vaswani, N.~Shazeer, N.~Parmar, J.~Uszkoreit, L.~Jones, A.~N. Gomez,
  {\L}.~Kaiser, and I.~Polosukhin.
\newblock Attention is all you need.
\newblock In {\em Advances in neural information processing systems}, pages
  5998--6008, 2017.

\bibitem{yu2014mixed}
D.~Yu, H.~Wang, P.~Chen, and Z.~Wei.
\newblock Mixed pooling for convolutional neural networks.
\newblock In {\em International Conference on Rough Sets and Knowledge
  Technology}, pages 364--375. Springer, 2014.

\bibitem{zhang2018learning}
M.~Zhang, Y.~Wu, W.~Li, and W.~Li.
\newblock Learning universal sentence representations with mean-max attention
  autoencoder.
\newblock {\em arXiv preprint arXiv:1809.06590}, 2018.

\bibitem{zhang2018graph}
S.~Zhang, H.~Tong, J.~Xu, and R.~Maciejewski.
\newblock Graph convolutional networks: Algorithms, applications and open
  challenges.
\newblock In {\em International Conference on Computational Social Networks},
  pages 79--91. Springer, 2018.

\end{thebibliography}
\bibliographystyle{abbrv}
\end{document}